\documentclass[runningheads]{llncs}
\usepackage{graphicx}
\usepackage[frozencache, cachedir=.]{minted}
\usepackage{todonotes}
\usepackage{hyperref}

\begin{document}
\title{German Tourism Knowledge Graph}
%
%
\author{Umutcan Serles\inst{1,2}\orcidID{0000-0001-6459-474X} \and
Elias K\"{a}rle\inst{2}\orcidID{0000-0002-2686-3221} \and
Richard Hunkel\inst{3} \and Dieter Fensel\inst{1}}

\authorrunning{U. Serles et al.}
%
\institute{STI Innsbruck, Department of Computer Science, University of Innsbruck \\
\email{\{name.surname\}@sti2.at} \and
Onlim GmbH \\
\email{\{name.surname\}@onlim.com}  \and
German National Tourist Board \\
\email{richard.hunkel@germany.travel}}
\maketitle              

\setcounter{footnote}{0}

\begin{abstract}

Tourism is one of the most critical sectors of the global economy. Due to its heterogeneous and fragmented nature, it provides one of the most suitable use cases for knowledge graphs. In this poster, we introduce the German Tourism Knowledge Graph that integrates tourism-related data from 16 federal states of Germany and various other sources to provide a curated knowledge source for various applications. It is publicly available through GUIs and an API.


\end{abstract}

\section{Introduction}

 Tourism is one of the most fragmented sectors, bringing many different services together, such as accommodation, transportation, events, and many more. From a data management perspective, such fragmentation and abundance of service providers bring a large size and heterogeneity in data sources. Most of the data are isolated behind closed systems, preventing digitalization in tourism from reaching its full potential. Using the power of integrated data and benefiting from the synergies between regional touristic offers can help overcome these obstacles.

With that motivation, the German National Tourist Board (GNTB)\footnote{In German, Deutsche Zentrale für Tourismus. \url{https://germany.travel}} commissioned the construction of German Tourism Knowledge Graph (GTKG) that integrates data from the regional marketing organizations\footnote{These are agencies that promote tourism in their regions} of 16 states in Germany. It was launched in May 2023 and being updated daily by the regional marketing organizations of German states. To the best of our knowledge, GTKG is the first tourism knowledge graph driven by stakeholders in the tourism industry that is continuously updated and maintained at a large scale.

In the remainder of this paper, we will first briefly describe different tasks for building the GTKG in Section \ref{sec:building-gtkg}. Then, we provide a literature review and comparison with similar knowledge graphs in Section \ref{sec:related-work}. We conclude the paper by summarizing the key points and indications for future work in Section \ref{sec:conclusion}.

\section{Building the German Tourism Knowledge Graph}
\label{sec:building-gtkg}
We will present the building process of GTKG in three steps, namely knowledge creation, enrichment and deployment.

\subsection{Knowledge Creation}

The knowledge creation task can be split in two: the creation of the schema (TBox) and the instantiation of the schema (ABox). The schema used in the GTKG is created by the Open Data Tourism Alliance \footnote{\url{https://open-data-germany.org/en/open-data-germany/}}. Open Data Tourism Alliance (ODTA) is an initiative consisting of tourism domain experts representing the DACH region (Germany, Austria and Switzerland). The main goal of the initiative is to create \textit{domain specifications} \cite{simsekDomainSpecificCustomizationSchema2020} for tourism-related data. Domain specifications are extended subsets of schema.org. ODTA selects the relevant subsets of types and properties from schema.org and extends them with new types and properties when schema.org is not enough for more detailed description of touristic items. Furthermore, constraints (e.g. cardinality) on the properties can be defined. The domain specifications are encoded as SHACL shapes\footnote{The list of domain specifications currently used by GTKG can be found here: \url{https://semantify.it/list/LRVOilZZ6}. There was a new release of ODTA domain specifications in February 2024 (\url{https://odta.sti2.org/}) and the migration process is still ongoing.}.

The regional marketing organizations of German states create instances of the types in the TBox and import it to GTKG on a daily basis. Each instance is verified against its corresponding SHACL shape (based on target declerations or explicit shape references on instances) to ensure that they satisfy all constraints provided by the domain specifications made by ODTA. The regional marketing organizations typically use RML\cite{dimouRMLGenericLanguage2014a} mappings between the ODTA schemas and the metadata of their internal systems.

The major types\footnote{See \url{https://semantify.it/list/LRVOilZZ6} for the domain specifications based on these types.} represented in GTKG (on 07/03/2024) are schema:Event (239K instances), odta:PointOfInterest (118K instances), odta:Trail (23K instances), schema:LodgingBusiness (25K instances), schema:LocalBusiness (70K instances) and schema:FoodEstablishment (20K instances)\footnote{\textit{schema} is the prefix for schema.org namespace. \textit{odta} is the prefix for https://odta.io/voc/ namespace that represents newly created types and properties. Instance counts are result of the query here: \url{https://purl.archive.org/gtkg/queries}}.

\subsection{Knowledge Enrichment}

The knowledge graph created based on the data provided by 16 regional marketing organizations is further enriched with data from other sources. A prominent example of enrichment is geo-linking of e-charging stations \footnote{\url{https://ladestationen.api.bund.dev/}} and public transportation stops \footnote{\url{https://www.opendata-oepnv.de/}} with points of interest. The enrichment process can be configured via the API by the data providers. The details of the configuration is presented in the API documentation (Section \ref{sec:deployment}) under "Guides". 

\subsection{Knowledge Deployment}
\label{sec:deployment}
The GTKG API\footnote{See \url{https://purl.archive.org/gtkg/api-doc} for a comprehensive documentation} enables programmatical access to the GTKG. It allows SPARQL queries against the graph but also many other ways of accessing the knowledge graph (e.g., search by name). The API needs an authentication key, which can be obtained from GNTB free of charge\footnote{\url{https://purl.archive.org/gtkg/access}}. Listing \ref{lst:api-call} shows an example request to the API to run the SPARQL query in \url{https://purl.archive.org/gtkg/queries}.

\vspace{-19pt}

\begin{listing}[h]
    \centering
\begin{minted}{shell}
curl --request POST \
     --url https://proxy.opendatagermany.io/api/ts/v1/kg/sparql \
     --header 'Accept: application/sparql-results+json' \
     --header 'X-API-KEY: fb0758867a9a01fce51546b0e537bb56' \
     --header 'content-type: text/plain' \
     --data 'PREFIX schema: <https://schema.org/>   PREFIX odta: <https://odta.io/voc/> SELECT ?s ?name ?poi (xsd:decimal(?dist) as ?distance) ?unit WHERE  { ?s a odta:PointOfInterest. ?s schema:name ?name. ?s odta:geoLink [odta:linkSource/schema:name ?poi; odta:walkingDistance/schema:value ?dist; odta:walkingDistance/schema:unitCode ?unit] } LIMIT 10'
\end{minted}
    \caption{A SPARQL query to return 10 results about POIs and the walking distances to the entities linked via geo-linking to them}
    \label{lst:api-call}
\end{listing}

\vspace{-19pt}
Additional to the API, the GTKG provides a graphical user interface. The Search Widget\footnote{Currently being tested on a subset of GTKG \url{https://purl.archive.org/gtkg/search}} allows text-based search of instances and filtering by type. Different visualization modalities can be accessed for an instance from the search results.

GTKG allows the regional marketing organizations to use their own instance identifiers, which may not be always dereferencable. To mitigate this drawback, GTKG provides a \textit{a wrapper URI} that takes the instance identifier as query parameter and creates an HTML depiction as well as provides a way of downloading an RDF serialization of the instance. See \url{https://purl.archive.org/gtkg/example} for an example.

\section{Related Work}
\label{sec:related-work}

The survey from Abu-Salih \cite{abu-salihDomainspecificKnowledgeGraphs2021} covers domain-specific knowledge graphs, including tourism-related knowledge graphs. These graphs typically target specific regions and may focus on specific kinds of tourism-related data. For example, Tyrol region in Austria \cite{karleBuildingEcosystemTyrolean2018a}, London in England, Madeira in Portugal and Cote d'Azur in France \cite{troncy3cixtyBuildingComprehensive2017}.

Although many other knowledge graphs exist in the tourism domain, the German Tourism Knowledge Graph stands out. It is not a purely academic project, but driven and funded by stakeholders in the tourism sector. It contains data about various tourism-related topics and has a national coverage instead of focusing a single region. It also provides various access modalities that can serve different purposes (e.g., machine or human consumption).

\section{Conclusion and Future Work}
\label{sec:conclusion}

We presented the German Tourism Knowledge Graph. The knowledge graph is a publicly accessible resource and uses a schema.org-based schema to describe the integrated data. It is important to note that the GTKG is built around standard semantic technologies like RDF(S), SPARQL, SHACL and widespread technologies like RML. Each instance in the knowledge graph has its own license, which is typically a Creative Common license like CC-BY-SA. 

The German National Tourism Board is currently internally testing a chatbot powered by the GTKG, which will be available for public soon. Many other applications have been developed at a hackathon in November 2023\footnote{\url{https://de.linkedin.com/posts/german-national-tourist-board\_kollaboration-innovation-dzt-activity-7132391151659016192-k5hW}}. Currently, individual instances can be downloaded in an RDF serialization. However, periodic releases of the entire knowledge graph are also considered in the future work.

%
%

\bibliography{references.bib}

\end{document}